\documentclass[11pt,letterpaper]{article}
\usepackage{emnlp2017}
\usepackage{times}
\usepackage{latexsym}
\usepackage{amsfonts} 
\usepackage[]{amsmath}
\usepackage[ruled,vlined]{algorithm2e}
\usepackage{algpseudocode}
\usepackage{graphicx}
\usepackage{multirow}
\usepackage{url}
\usepackage{colortbl}
\usepackage{paralist}
\usepackage[utf8]{inputenc}
\usepackage{cleveref}
\crefname{section}{§}{§§}
\Crefname{section}{§}{§§}
\DeclareMathOperator*{\argmax}{arg\,max}

\emnlpfinalcopy

\def\emnlppaperid{***}

\title{MUSE: Modularizing Unsupervised Sense Embeddings}

 \author{Guang-He Lee$^\star$$^\dag$ \quad Yun-Nung Chen$^\dag$\\$^\star$Massachusetts Institute of Technology, Cambridge, MA\\$^\dag$National Taiwan University, Taipei, Taiwan \\ {\tt guanghe@mit.edu\quad y.v.chen@ieee.org}
         }
\date{}

\begin{document}

\maketitle

\begin{abstract}
% No change from draft
This paper proposes to address the word sense ambiguity issue in an unsupervised manner, where word sense representations are learned along a word sense selection mechanism given contexts. Prior work focused on designing a single model to deliver both mechanisms, and thus suffered from either coarse-grained representation learning or inefficient sense selection. The proposed modular approach, \emph{MUSE}, implements flexible modules to optimize distinct mechanisms, achieving the first purely sense-level representation learning system with linear-time sense selection. We leverage reinforcement learning to enable joint training on the proposed modules, and introduce various exploration techniques on sense selection for better robustness. The experiments on benchmark data show that the proposed approach achieves the state-of-the-art performance on synonym selection as well as on contextual word similarities in terms of MaxSimC.
\end{abstract}

\section{Introduction}

Recently, deep learning methodologies have dominated several research areas in natural language processing (NLP), such as machine translation, language understanding, and dialogue systems.
However, most of applications usually utilize word-level embeddings to obtain semantics.
Considering that natural language is highly ambiguous, the standard word embeddings may suffer from polysemy issues.
% the original argument:
\newcite{neelakantan2015efficient} pointed out that, due to triangle inequality in vector space, if one word has two different senses but is restricted to \emph{one embedding}, the sum of the distances between the word and its synonym in each sense would upper-bound the distance between the respective synonyms, which may be mutually irrelevant, in embedding space\footnote{$d(\text{rock},\text{stone})+d(\text{rock},\text{shake}) \geq d(\text{stone}, \text{shake})$}.
%however, the semantically similar words in each sense may be actually semantically irrelevant.
Due to the theoretical inability to account for polysemy using a single embedding representation per word, multi-sense word representations are proposed to address the ambiguity issue using multiple embedding representations for different senses in a word~\cite{reisinger2010multi,HuangEtAl2012}.

This paper focuses on unsupervised learning from the unannotated corpus. 
There are two key mechanisms for a multi-sense word representation system in such scenario: 
1) a sense selection (decoding) mechanism infers the most probable sense for a word given its context and 
2) a sense representation mechanism learns to embed word senses in a continuous space.

Under this framework, prior work focused on designing a single model to deliver both mechanisms \cite{neelakantan2015efficient,li2015multi,Qiu2016ContextDependentSE}. 
However, the previously proposed models introduce side-effects:
%the single model suffers from modeling difficulty that efficient sense selection comes at the cost of ambiguity in the sense representation learning procedure, vice versa.
%Existing side-effects include 
1) mixing word-level and sense-level tokens achieves efficient sense selection but introduces ambiguous word-level tokens during the representation learning process~\cite{neelakantan2015efficient,li2015multi}, and 2) pure sense-level tokens prevent ambiguity from word-level tokens but require exponential time complexity when decoding a sense sequence \cite{Qiu2016ContextDependentSE}.

Unlike the prior work, this paper proposes \emph{MUSE}\footnote{The trained models and code are available at \url{https://github.com/MiuLab/MUSE}.}---a novel modularization framework incorporating sense selection and representation learning models, which implements flexible modules to optimize distinct mechanisms. 
Specifically, MUSE enables linear time sense identity decoding with a \emph{sense selection module} and purely sense-level representation learning with a \emph{sense representation module}.

With the modular design, we propose a novel joint learning algorithm on the modules by connecting to a reinforcement learning scenario, which achieves the following advantages. 
First, the decision making process under reinforcement learning better captures the sense selection mechanism than probabilistic and clustering methods. 
Second, our reinforcement learning algorithm realizes the first single objective function for modular unsupervised sense representation systems. 
Finally, we introduce various exploration techniques under reinforcement learning on sense selection to enhance robustness.

In summary, our contributions are five-fold:
\begin{compactitem}
  \item \emph{MUSE} is the first system that maintains purely sense-level representation learning with linear-time sense decoding.
  \item We are among the first to leverage reinforcement learning to model the sense selection process in sense representations system.
  \item We are among the first to propose a single objective for modularized unsupervised sense embedding learning.
  \item We introduce a sense exploration mechanism for the sense selection module to achieve better flexibility and robustness.
  \item Our experimental results show the state-of-the-art performance for synonym selection and contextual word similarities in terms of MaxSimC.
\end{compactitem}

\section{Related Work}

There are three dominant types of approaches for learning multi-sense word representations in the literature: 1) clustering methods,  2) probabilistic modeling methods, and 3) lexical ontology based methods.
Our reinforcement learning based approach can be loosely connected to clustering methods and probabilistic modeling methods.

\citet{reisinger2010multi} first proposed multi-sense word representations on the vector space based on clustering techniques. 
With the power of deep learning, some work exploited neural networks to learn embeddings with sense selection based on clustering~\cite{HuangEtAl2012,neelakantan2015efficient}.
\citet{chen2014unified} replaced the clustering procedure with a word sense disambiguation model using WordNet~\cite{miller1995wordnet}.
\citet{kaageback2015neural} and \citet{vu2016k} further leveraged a weighting mechanism and interactive process in the clustering procedure.
Moreover, \citet{guo2014learning} leveraged bilingual resources for clustering.
However, most of the above approaches separated the clustering procedure and the representation learning procedure without a joint objective, which may suffer from the error propagation issue.
Instead, the proposed approach, MUSE, enables joint training on sense selection and representation learning.

Instead of clustering, probabilistic modeling methods have been applied for learning multi-sense embeddings in order to make the sense selection more flexible,
where \citet{tian2014probabilistic} and \citet{jauhar2015ontologically} conducted probabilistic modeling with EM training.
\citet{li2015multi} exploited Chinese Restaurant Process to infer the sense identity. %and demonstrates efficacy of multi-sense word representations on several downstream NLP tasks.
Furthermore, \citet{bartunov2015breaking} developed a non-parametric Bayesian extension on the skip-gram model~\cite{mikolov2013distributed}. % for multi-sense embeddings.
Despite reasonable modeling on sense selection, all above methods mixed word-level and sense-level tokens during representation learning---unable to conduct representation learning in the pure sense level due to the complicated computation in their EM algorithms. %~\cite{bartunov2015breaking}. 

Recently, \citet{Qiu2016ContextDependentSE} proposed an EM algorithm to learn purely sense-level representations, where the computational cost is high when decoding the sense identity sequence, because it takes exponential time to search all sense combination within a context window.
Our modular design addresses such drawback, where the sense selection module decodes a sense sequence with linear-time complexity, while the sense representation module remains representation learning in the pure sense level.

%Another difference between the prior work and ours is that, \citet{Qiu2016ContextDependentSE} used WordNet~\cite{miller1995wordnet} to obtain the number of senses for words, and our solution supports non-parametric learning for automatically deciding the sense number of each word.

\begin{figure*}
\centering
  \includegraphics[width=1\linewidth]{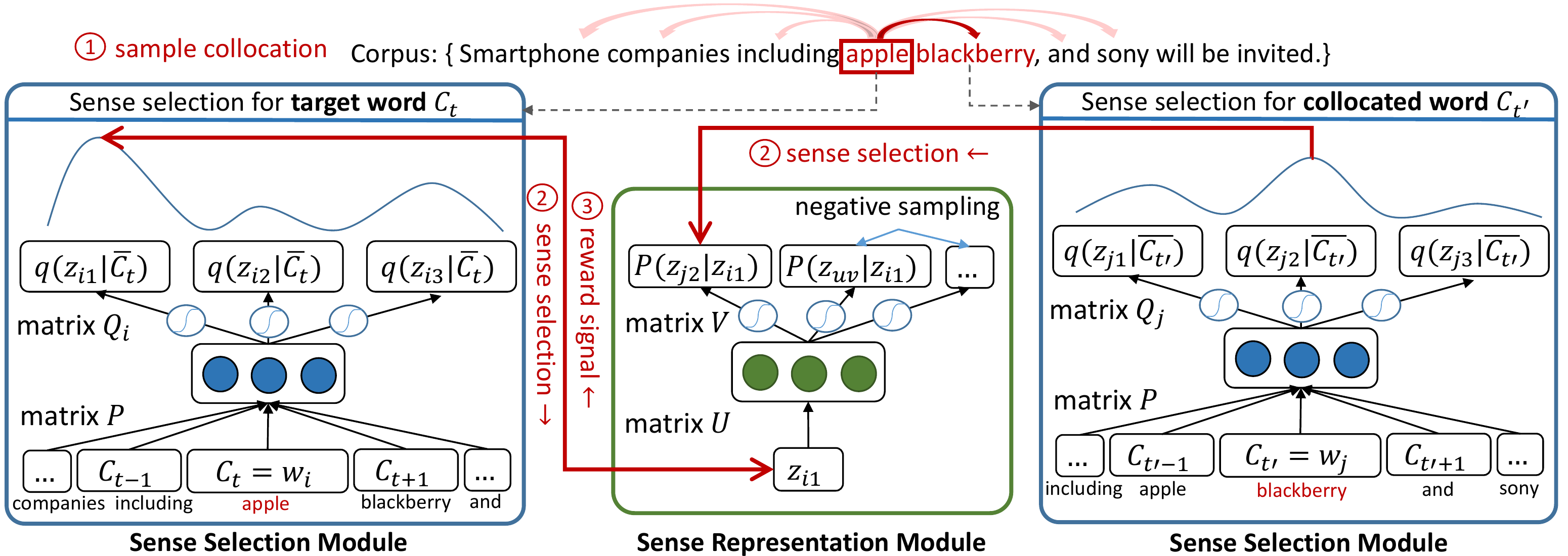}
  \vspace{-4mm}
  \caption{The \emph{MUSE} architecture with a 3-step learning algorithm: 1) collocation sampling, 2) sense selection for sense representation learning, and 3) optimizing sense selection with a reward signal from sense representation. Reward signal is only passed to the target word to stabilize model training due to directional architecture in the sense representation module.}
  \label{fig:model}
  \vspace{-2mm}
\end{figure*}

%Furthermore, most related work either exploited sense representations directly~\cite{neelakantan2015efficient,Qiu2016ContextDependentSE} or in combination with online expectation sense counts~\cite{li2015multi,jauhar2015ontologically} to conduct sense selection. The proposed approach incorporates a distinct sense selection module for modularization. %, which can be learned together with the representation learning module through reinforcement learning.

Unlike a lot of relevant work that requires additional resources such as the lexical ontology~\cite{pilehvar2016conflated,rothe2015autoextend, jauhar2015ontologically,chen2015improving,iacobacci2015sensembed} or bilingual data~\cite{guo2014learning,ettinger2016retrofitting,vsuster2016bilingual},
which may be unavailable in some language, 
our model can be trained using only an unlabeled corpus.
Also, some prior work proposed to learn topical embeddings and word embeddings jointly in order to consider the contexts~\cite{liu2015learning,liu2015topical}, whereas this paper focuses on learning multi-sense word embeddings.

%This paper combines the benefits of clustering and probabilistic modeling to propose a joint learning framework in a reinforcement learning manner without the need of knowledge basis and predefined sense numbers.

%, but the computational cost on decoding sense identity sequences is very high, $O(N^{4m}l+(N+m)N^{2m}lD)$, where $N$ is the maximal sense number per word, $l$ is the sentence length, $m$ is the context window size, and $D$ is embedding dimension. Instead, our model strikes a balance between computationally expensive sense decoding and pure sense-level representation, by adopting separate modules for sense selection and sense representation, which achieve linearly independent sense decoding ($O((N+m)lD)$), which will be derived in Section 3.2. Finally, \cite{Qiu2016ContextDependentSE} use WordNet~\cite{miller1995wordnet} to obtain the number of senses for words, while our solution supports non-parametric learning.

\section{Proposed Approach: MUSE}

This work proposes a framework to modularize two key mechanisms for multi-sense word representations: a \emph{sense selection module} and a \emph{sense representation module}. 
The sense selection module decides which sense to use given a text context, whereas the sense representation module learns meaningful representations based on its statistical characteristics.
%Unlike prior work that must compromise between efficient sense selection and purely sense-level representation learning,
Unlike prior work that must suffer from either inefficient sense selection~\cite{Qiu2016ContextDependentSE} or coarse-grained representation learning~\cite{neelakantan2015efficient,li2015multi,bartunov2015breaking},
the proposed modularized framework is capable of performing efficient sense selection and learning representations in pure sense level simultaneously.

To learn sense-level representations, a sense selection model should be first established for sense identity decoding.
On the other hand, the sense embeddings should guide the sense selection model when decoding a sense identity sequence.
Therefore, these two modules should be tangled.
This indicates that a naive two-stage algorithm or two separate learning algorithms proposed by prior work are not optimal.

By connecting the proposed formulation with reinforcement learning literature, we design a novel joint training algorithm. Besides, taking advantage of the form of reinforcement learning, we are among the first to investigate various exploration techniques in sense selection for unsupervised sense embedding learning. 

\subsection{Model Architecture}
Our model architecture is illustrated in Figure~\ref{fig:model}, where there are two modules in optimization.

\subsubsection{Sense Selection Module}
Formally speaking, given a corpus $C$, vocabulary $W$, and the $t$-th word $C_t=w_i \in W$, we would like to find the most probable sense $z_{ik} \in Z_i$, where $Z_i$ is the set of senses in word $w_i$.
Assuming that a word sense is determined by the local context, we exploit a local context $\bar{C_t} = \{C_{t-m},\cdots,C_{t+m}\}$ for sense selection according to the Markov assumption, where $m$ is the size of a context window. 
Then we can either formulate a probabilistic policy $\pi(z_{ik} \mid \bar{C_t})$ about sense selection or estimate the \emph{individual} likelihood $q(z_{ik} \mid \bar{C_t})$ for each sense identity.

To ensure efficiency, here we exploit a linear neural architecture that takes word-level input tokens and outputs sense-level identities. 
The architecture is similar to continuous bag-of-words (CBOW)~\cite{mikolov2013efficient}.
Specifically, given a \textit{word} embedding matrix $P$, the local context can be modeled as the summation of word embeddings from its context $\bar{C_t}$. 
The output can be formulated with a 3-mode tensor $Q$, whose dimensions denote words, senses, and latent variables.
Then we can model $\pi(z_{ik} \mid \bar{C_t})$ or $q(z_{ik} \mid \bar{C_t})$ correspondingly.
Here we model $\pi(\cdot)$ as a \emph{categorical} distribution using a softmax layer:
\begin{equation}
	%\vspace{-5pt}
    \pi(z_{ik} \mid \bar{C_t}) = \frac{\exp(Q_{ik}^T\sum_{j\in \bar{C_t}}P_j)}{\sum_{k'\in Z_i} \exp(Q_{ik'}^T\sum_{j\in \bar{C_t}}P_j)}.
    \label{eq:pg}
   	%\vspace{-3pt}
\end{equation}
On the other hand, the likelihood of selecting \emph{distinct} sense identities, $q(z_{ik}\mid \bar{C_t})$, is modeled as a \emph{Bernoulli} distribution with a sigmoid function $\sigma(\cdot)$:
\begin{equation}
    q(z_{ik} \mid \bar{C_t}) = \sigma( Q_{ik}^T\sum_{j \in \bar{C_t}}P_j ).
    \label{eq:q}
   	%\vspace{-6pt}
\end{equation}
Different modeling approaches require different learning methods,
%The choice of modeling affects the choice of learning method,
especially for the unsupervised setting. % where the reliable supervision is unavailable. 
We leave the corresponding learning algorithms in \cref{sec:rl}.
%Here we only list possible modeling settings for the sense selection module and leave the learning algorithm on \cref{sec:rl} when the sense representation module is also presented.
Finally, with a built sense selection module, we can apply any selection algorithm such as a greedy selection strategy to infer the sense identity $z_{ik}$ given a word $w_i$ with its context $C_t$.

We note that modularized model enables efficient sense selection by leveraging word-level tokens, while remaining purely sense-level tokens in the representation module.
Specifically, if $n$ denotes $\max_k |Z_k|$, decoding $L$ words takes $O(nL)$ senses to be searched due to independent sense selection. 
The prior work using a single model with purely sense-level tokens~\cite{Qiu2016ContextDependentSE} requires exponential time to calculate the collocation energy for every possible combination of sense identities within a context window, $O(n^{2m})$, for a \emph{single target sense}. Further, \newcite{Qiu2016ContextDependentSE} took an additional sequence decoding step with quadratic time complexity $O(n^{4m}L)$, based on an exponential number $n^{2m}$ in the base unit.
It demonstrates the achievement about sense inference efficiency in our proposed model.
%We will provide training details and the formulation for $q_\theta(z_{ik}|\bar{C_t})$ on \cref{sec:rl}. As long we have some probabilistic policy or fitness estimation, we can simply employ a greedy strategy to select sense, whose details and discussions will also be available on \cref{sec:rl}.

%In addition, assuming that a word sense is determined by the local context, we use the Markov property to formulate the sense selection module as a Markov Decision Process (MDP) to infer the most probable sense based on its local context.

%Assuming that a word sense is determined by the local context, we use the Markov property to formulate the sense selection module as a Markov Decision Process (MDP) to infer the most probable sense based on its local context.
%Given a corpus $C$ and a context window $m$, we extract a local context $\bar{C_t} = \{C_{t-m},...,C_{t-1},C_{t+1},...,C_{t+m}\}$ as a state $S_t$, and the selection of a sense $z_{ik} \in Z_i$ for a word $w_i=C_t$ given context $\bar{C_t}$ as an action $A_{ik}$ given the state $S_t$, where $Z_i$ is the set of senses of word $w_i$. % when the specific sense $z_{ik}$ of a word $w_i$ can be inferred given the context $\bar{C_t}$.
%However, this formulation lacks a central element of MDP: a reward signal for measuring the fitness of the selected sense $z_{ik}$.
%Therefore, if we have meaningful sense representations containing statistical estimation, we can use the estimation as a surrogate reward to simulate the reward signal in the sense selection module.

\subsubsection{Sense Representation Module}

%A successful sense selection module can be applied to each word in a corpus for obtaining its sense identity.
%Various techniques about word embeddings can be directly employed after mapping all words in a corpus to its sense identity. % and then learning meaningful sense representations.

%Once sense identity is available, we can replace each word identity with its sense identity in a corpus as the learning target to train a semantic representation using existing representation learning algorithms. 

The semantic representation learning is typically formulated as a maximum likelihood estimation (MLE) problem for collocation likelihood.
%The typical method for learning semantic representations is to formulate the learning problem as a maximum likelihood estimation (MLE) problem for collocation likelihood. 
%However, for unsupervised sense representation learning, sense selection should be first performed to obtain the sense identity for representation learning, so it is desirable to include only a few sense instances in each stochastic update for representation learning. 
In this paper, we use the skip-gram formulation~\cite{mikolov2013distributed} considering that it requires less training time,
%because it is more light-weight and thus requires less training time:
where only two sense identities are required for stochastic training.
Other popular candidates, like GloVe~\cite{pennington2014glove} and CBOW~\cite{mikolov2013efficient}, require more sense identities to be selected as input and thus not suitable for our scenario. % that sense selections should be performed prior to representation learning. 
For example, GloVe~\cite{pennington2014glove} takes computationally expensive collocation \emph{counting} statistics for each token in a corpus as input, which requires sense selection for every occurrence of the target word across the whole corpus for a single optimization step.

To learn the representations, we first create input sense representation matrix $U$ and collocation estimation matrix $V$ as the learning targets.
Given a target word $w_i$ and collocated word $w_j$ with corresponding local contexts, we map them to their sense identities as $z_{ik}$ and $z_{jl}$ by the sense selection module, and maximize the sense collocation log likelihood $\log \mathcal{L}(\cdot)$. A natural choice of the likelihood function is formulated as a categorical distribution over all possible collocated senses given the target sense $z_{ik}$:
\begin{equation}
\label{eq:categorical}
	 \max_{U,V}\;\;\log \mathcal{L}(z_{jl} \mid z_{ik}) = \log \frac{\exp (U_{z_{ik}}^TV_{z_{jl}})}{\sum_{z_{uv}}\exp(U_{z_{ik}}^TV_{z_{uv}})}.
\end{equation}
Instead of enumerating all possible collocated senses which is computationally expensive, we use the skip-gram objective (\ref{w2v})~\cite{mikolov2013distributed} to approximate (\ref{eq:categorical}) as shown in the green block of Figure~\ref{fig:model}.
%\vspace{-4pt}
\begin{align}
\label{w2v}
	 \max_{U,V}\;\;\log &\, \bar{\mathcal{L}}(z_{jl} \mid  z_{ik}) = \log \sigma(U_{z_{ik}}^TV_{z_{jl}}) \\
    & + \sum_{v=1}^M \mathbb{E}_{z_{uv}\sim p_{neg}(z)}[\log \sigma(-U_{z_{ik}}^T V_{z_{uv}})],\nonumber
    \vspace{-3pt}
\end{align}
%\vspace{-4pt}
where $p_{neg}(z)$ is the distribution over all senses for negative samples. In our experiment with $|Z_i|$ senses for word $w_i$, we use $(1/{|Z_i|})$ word-level unigram as sense-level unigram for efficiency and the $3/4$-th power trick in \citet{mikolov2013distributed}.

We note that our modular framework can easily maintain purely sense-level tokens with an arbitrary representation learning model. 
In contrast, most related work using probabilistic modeling \cite{tian2014probabilistic,jauhar2015ontologically,li2015multi,bartunov2015breaking} binded sense representations with the sense selection mechanism, so efficient sense selection by leveraging word-level tokens can be achieved only at the cost of mixing word-level and sense-level tokens in their representation learning process.
%binded the sense selection mechanism with sense representations, so they can perform efficient sense selection but suffer from embedding ambiguity due to mixing word-level and sense-level tokens in the representation learning process.

%bind their sense selection mechanism with their sense representations, thus achieving efficient sense selection like our model architecture at the cost of mixing word-level and sense-level tokens in their representation learning process.

\subsection{Learning}
\label{sec:rl}

Without the supervised signal for the proposed modules, it is desirable to connect two modules in a way where they can improve each other by their own estimations. 
First, a trivial way is to forward the prediction of the sense selection module to the representation module.
Then we cast the estimated collocation likelihood as a reward signal for the selected sense for effective learning.
%Based on different modeling methods ((\ref{eq:pg}) or (\ref{eq:q})) in the sense selection module, we connect the model to respective learning algorithms. 
 
To realize the above procedure, we cast the learning problem a one-step Markov decision process (MDP) \cite{sutton1998reinforcement}, where the state, action, and reward correspond to context $\bar{C_t}$, sense $z_{ik}$, and collocation log likelihood $\log \bar{\mathcal{L}}(\cdot)$, respectively.
%The intuition about formulating as reinforcement learning is that, we do not know how many senses each word has
%Note that we focus on stochastic optimization for a practical learning setting.
Based on different modeling methods ((\ref{eq:pg}) or (\ref{eq:q})) in the sense selection module, we connect the model to respective reinforcement learning algorithms to solve the MDP. 
Specifically, we refer (\ref{eq:pg}) to policy distribution and refer (\ref{eq:q}) to Q-value estimation in the reinforcement learning literature.

The proposed MDP framework embodies several nuances of sense selection. First, the decision of a word sense is Markov: taking the whole corpus into consideration is not more helpful than a handful of necessary local contexts. Second, the decision making in MDP exploits a hard decision for selecting sense identity, which captures the sense selection process more naturally than a joint probability distribution among senses~\cite{Qiu2016ContextDependentSE}. % over possible sense identities. 
Finally, we exploit the reward mechanism in MDP to enable joint training: the estimation of sense representation is treated as a reward signal to guide sense selection. In contrast, the decision making under clustering~\cite{HuangEtAl2012,neelakantan2015efficient} considers the similarity within clusters instead of the outcome of a decision using a reward signal as MDP.

\subsubsection{Policy Gradient Method}
\label{sec:pg}

Because (\ref{eq:pg}) fits a valid probability distribution, an intuitive optimization target is the expectation of resulting collocation likelihood among each sense. %That says, an optimal policy $\pi(\cdot)$ in (\ref{eq:pg}) should leads to an optimal collocation likelihood in expectation. 
In addition, as the skip-gram formulation in (\ref{w2v}) is unidirectional ($\bar{\mathcal{L}}(z_{ik} \mid z_{jl}) \neq \bar{\mathcal{L}}(z_{jl} \mid z_{ik})$), we perform one-side optimization for the target sense $z_{ik}$ to stabilize model training\footnote{We observe about $4\%$ performance drop by optimizing input selection $z_{ik}$ and output selection $z_{jl}$ simultaneously.}. 
That is, for the target word $w_i$ and the collocated word $w_j$ given respective contexts $\bar{C_t}$ and $\bar{C_{t'}}$ ($0<|t-t'|\leq m$), we first draw a sense $z_{jl}$ for $w_j$ from the policy $\pi(\cdot \mid \bar{C_{t'}})$ and optimize the expected collocation likelihood for the target sense $z_{ik}$ as follows,
\begin{equation}
\label{policy_grad}
\max_{P,Q}\;\; \mathbb{E}_{z_{ik} \sim \pi(\cdot \mid \bar{C_t})}[\log \bar{\mathcal{L}}(z_{jl} \mid z_{ik})].
\end{equation}
Note that (\ref{w2v}) can be merged into (\ref{policy_grad}) as a single objective.
The objective is differentiable and supports stochastic optimization \cite{lei2016rationalizing}, which uses a stochastic sample $z_{ik}$ for optimization. 

%In terms of sense selection, we use sampling to approximate the expectation during training as $\pi(\cdot)$ fits a probability distribution, while we select the most likely sense during testing. 

%We refer readers to Appendix for detailed derivation of the policy gradient for (\ref{policy_grad}).

However, there are two possible disadvantages in this formulation. 
First, because the policy assumes the probability distribution in (\ref{eq:pg}), optimizing the selected sense must affect the estimation of the other senses.
%the selected senses must indirectly affect the estimation of other senses during optimization.
%updating it using the selected sense must affect the estimation for the other senses whose collocation likelihood is not involved during optimization. 
Second, if applying stochastic gradient ascent to optimizing (\ref{policy_grad}), it would always lower the probability estimation for the selected sense $z_{ik}$ even if the model accurately selects the right sense.
The detailed proof is in Appendix \ref{doubly_sg}.

\subsubsection{Value-Based Method} %Q-Learning} %and Reward Passing}
To address the above issues, we apply the Q-learning algorithm~\cite{mnih2013playing}. % with a novel reward passing procedure to this task.
Instead of maintaining a probabilistic policy for sense selection, Q-learning estimates the Q-value (resulting collocation log likelihood) for each sense candidate directly and independently.
Thus, the estimation of unselected senses may not be influenced by the selected one.
Note that in one-step MDP, the reward is equivalent to the Q-value, so we will use reward and Q-value interchangeably, hereinafter, based on the context.
%In this work, we correspond the Q-value to the collocation likelihood in (\ref{w2v}) and the estimated Q-value to (\ref{eq:q}).
%the estimated Q-value in this work to equation (\ref{eq:q}) and the estimated target as .

We further follow the convention of recent neural reinforcement learning by reducing the reward range to aid model training~\cite{mnih2013playing}. 
Specifically, we replace the \emph{log likelihood} $\log \bar{\mathcal{L}}(\cdot)\in (-\inf, 0]$ with the \emph{likelihood} $\bar{\mathcal{L}}(\cdot)\in [0,1]$ as the reward function.
Due to the monotonic operation in $\log()$, the relative ordering of the reward remains the same.

Furthermore, we exploit the probabilistic nature of likelihood for Q-learning. 
To elaborate, as Q-learning is used to approximate the Q-value for each action in typical reinforcement learning, most literature adopted square loss to characterize the discrepancy between the target and estimated Q-values~\cite{mnih2013playing}. 
In our setting where the Q-value/reward is a likelihood function, our model exploits cross-entropy loss to better capture the characteristics of probability distribution. 

Given that the collocation likelihood in (\ref{w2v}) is an \emph{approximation} to the original categorical distribution with a softmax function shown in (\ref{eq:categorical}) \cite{mikolov2013distributed}, we revise the formulation by omitting the negative sampling term. The resulting formulation $\hat{\mathcal{L}}(\cdot)$ is a Bernoulli distribution indicating whether $z_{jl}$ collocates or not given $z_{ik}$:
\begin{equation}
\hat{\mathcal{L}}(z_{jl} \mid z_{ik}) = \sigma(U^T_{z_{ik}}V_{z_{jl}}).
\end{equation}
There are three advantages about using $\hat{\mathcal{L}}(\cdot)$ instead of approximated $\bar{\mathcal{L}}(\cdot)$ and original $\mathcal{L}(\cdot)$. First, regarding the variance of estimation, $\hat{\mathcal{L}}(\cdot)$ better captures $\mathcal{L}(\cdot)$ than $\bar{\mathcal{L}}(\cdot)$ because $\bar{\mathcal{L}}(\cdot)$ involves sampling:
%$\bar{\mathcal{L}}(\cdot)$ involves sampling and thus has positive variance, while $\hat{\mathcal{L}}(\cdot)$ and $\mathcal{L}(\cdot)$ have no variance at all.
\begin{equation}
Var(\bar{\mathcal{L}}(\cdot)) \geq Var(\hat{\mathcal{L}}(\cdot)) = Var(\mathcal{L}(\cdot)) = 0.
\end{equation}
Second, regarding the relative ordering of estimation, for any two collocated senses $z_{jl}$ and $z_{jl'}$ with a target sense $z_{ik}$, the following equivalence holds:
\begin{eqnarray}
&& \mathcal{L}(z_{jl} \mid z_{ik}) < \mathcal{L}(z_{jl'} \mid z_{ik})\\\nonumber
&\Leftrightarrow & \bar{\mathcal{L}}(z_{jl} \mid z_{ik}) < \bar{\mathcal{L}}(z_{jl'} \mid z_{ik})\\\nonumber
&\Leftrightarrow & \hat{\mathcal{L}}(z_{jl} \mid z_{ik}) < \hat{\mathcal{L}}(z_{jl'} \mid z_{ik})\nonumber
\end{eqnarray}
Third, for collocation computation, $\mathcal{L}(\cdot)$ requires all sense identities and $\bar{\mathcal{L}}(\cdot)$ requires $(M+1)$ sense identities, whereas $\hat{\mathcal{L}}(\cdot)$ only requires $1$ sense identity. 
In sum, the proposed $\hat{\mathcal{L}}(\cdot)$ approximates $\mathcal{L}(\cdot)$ with no variance, no ``bias'' (in terms of relative ordering), and significantly less computation.
%the likelihood estimation in (\ref{w2v}) as a Bernoulli distribution indicating whether $z_{ik}$ and $z_{jl}$ collocated or not. In next section, we will show that we can simply omit the negative sampling terms in (\ref{w2v}) for a more robust estimation.

Finally, because both target distribution $\hat{\mathcal{L}}(\cdot)$ and estimated distribution $q(\cdot)$ in (\ref{eq:q}) are Bernoulli distributions, we follow the last section to conduct one-side optimization by fixing a collocated sense $z_{jl}$ and optimize the selected sense $z_{ik}$ with cross entropy as
\begin{equation}
%\vspace{-3mm}
\label{Qlearn}
 \min_{P,Q} \;\; H(\hat{\mathcal{L}}(z_{ik}\mid z_{jl}), q(z_{ik}\mid \bar{C_{t}})).%\\
%= & \min \;\;-\hat{\mathcal{L}}(z_{ik}\mid z_{jl}) \log q(z_{ik}\mid \bar{C_{t}})\nonumber\\
% & -(1-\hat{\mathcal{L}}(z_{ik}\mid z_{jl})) \log (1-q(z_{ik}\mid \bar{C_{t}})).\nonumber
\end{equation}

\subsubsection{Joint Training}

To jointly train sense selection and sense representation modules, we first select a pair of the collocated senses, $z_{ik}$ and $z_{jl}$, based on the sense selection module with any selecting strategy (e.g. greedy), and then optimize the sense representation module and the sense selection module using the above derivations. Algorithm~\ref{algo} describes the proposed MUSE model training procedure.

As modular frameworks, the major distinction between our modular framework and two-stage clustering-representation learning framework \cite{neelakantan2015efficient,vu2016k} is that we establish a reward signal from the sense representation to the sense selection module to enable immediate and joint optimization.
%Finally, we note that (\ref{w2v}) can be merged into (\ref{eq:pg}) as a single objective.

%To jointly train sense selection and sense representation modules, we first sample a collocated word pair $w_i, w_j$ with respective contexts $\bar{C_t},\bar{C_{t'}}$, and use the estimated Q-value to select the most probable senses $z_{ik}, z_{jl}$.
%The selected senses are passed to the sense representation module to optimize the sense collocation likelihood.
%Afterwards, the estimated collocation likelihood is passed back as a reward signal to optimize the sense selection module.
%The loss function is defined as cross-entropy $H(\cdot)$ due to the probability distribution.

%There are two major contributions in our modular design.
%First, efficient sense selection with word embeddings and pure sense representation learning are simultaneously achieved.
%Second, reinforcement learning allows both modules to be jointly trained.

%Specifically, since the reward signal here is presented as log-likelihood $\mathcal{L}(\cdot)$, we conduct the range reduction by its characteristics.
%Specifically, we use the collocation \emph{likelihood} instead of its monotonous \emph{log-likelihood} because likelihood is bounded within $[0,1]$. 

\begin{algorithm}[t]
\caption{Learning Algorithm}
\label{algo}
 \For{$w_i=C_t\in C$}{
   sample $w_j=C_{t'} (0<|t'-t|\leq m$)\;
   $z_{ik}=$ select$({C_t}, w_i)$\;
   $z_{jl}=$ select$({C_{t'}}, w_j)$\;
   optimize $U,V$ by (\ref{w2v}) for the sense representation module\;
   optimize $P,Q$ by (\ref{policy_grad}) or (\ref{Qlearn}) for the sense selection module\;
 }
\end{algorithm}

\subsection{Sense Selection Strategy}

Given a fitness estimation for each sense, exploiting the greedy sense is the most popular strategy for clustering algorithms \cite{neelakantan2015efficient, kaageback2015neural} and hard-EM algorithms \cite{Qiu2016ContextDependentSE, jauhar2015ontologically} in literature. However, there are two incentives to conduct exploration. First, in the early training stage when the fitness is not well estimated, it is desirable to explore underestimated senses. Second, due to high ambiguity in natural language, sometimes multiple senses in a word would fit in the same context. The dilemma between exploring sub-optimal choices and exploiting the optimal choice is called exploration-exploitation trade-off in reinforcement learning \cite{sutton1998reinforcement}.

We introduce exploration mechanisms for sense selection for both policy gradient and Q-learning.
For policy gradient, we sample the policy distribution to approximate the expectation in (\ref{policy_grad}).
Because of the flexible formulation of Q-learning, the following classic exploration mechanisms are applied to sense selection:
\begin{compactitem}
  \item \emph{Greedy}: selects the sense with the largest Q-value (no exploration).
  \item \emph{$\epsilon$-Greedy}: selects a random sense with $\epsilon$ probability, and adopts the greedy strategy otherwise \cite{mnih2013playing}.
%  \item \emph{Thompson}: select the sense with the largest Q-value estimated under dropout~\cite{srivastava2014dropout,Gal2016Dropout}.
  \item \emph{Boltzmann}: samples the sense based on the Boltzmann distribution modeled by Q-value. We directly use (\ref{eq:pg}) as the Boltzmann distribution for simplicity. 
\end{compactitem}
We note that Q-learning with Boltzmann sampling yields the same sampling process as policy gradient but different optimization objectives. 
To our best knowledge, we are among the first to explore several exploration strategies for unsupervised sense embedding learning.

In the following sections, MUSE-Policy denotes the proposed MUSE model with policy learning and MUSE-Greedy denotes the model using corresponding sense selection strategy for Q-learning.
%respective sense selection strategies for Q-learning, e.g., MUSE - Greedy, and denote the policy gradient method as MUSE - Policy.
%, despite that the Boltzmann distribution is exactly the one used in policy gradient, it still conduct independent optimization using Q-learning. The above strategies are possible selection strategies in Algorithm~\ref{algo}.  

%Due to high ambiguity in natural language, a greedy sense selection strategy may not work well in the early training stage, because the sense selection module does not learn well.
%Such a drawback also exists in the literature using clustering algorithms \cite{neelakantan2015efficient, kaageback2015neural} and hard-EM algorithms \cite{Qiu2016ContextDependentSE, jauhar2015ontologically}.
%This issue is known as exploration-exploitation trade-off \cite{sutton1998reinforcement}.

%\newcite{li2015multi} proposed to sample on a Chinese Restaurant Process to introduce uncertainty.
%In our neural network architecture for computing Q-values, we perform dropout~\cite{srivastava2014dropout} in the hidden layer $\sum_{j\in \bar{C_t}} P_j$ for our sense selection module to introduce uncertainty.

%However, as we adopt a neural network architecture to infer the Q-value, we can simply add dropout 

\section{Experiments}

We evaluate our proposed MUSE model in both quantitative and qualitative experiments. 

\subsection{Experimental Setup}
Our model is trained on the April 2010 Wikipedia dump~\cite{wikidump}, which contains approximately 1 billion tokens.
For fair comparison, we adopt the same vocabulary set as \citet{HuangEtAl2012} and \citet{neelakantan2015efficient}.
For preprocessing, we convert all words to their lower cases, apply the Stanford tokenizer and the Stanford sentence tokenizer \cite{manning-EtAl:2014:P14-5}, and remove all sentences with less than 10 tokens. 
The number of senses per word in $Q$ is set to 3 as the prior work~\cite{neelakantan2015efficient}. 
%We refer implementation details to Appendix \ref{parameter_setting}.

In the experiments, the context window size is set to 5 ($|\bar{C_t}|=11$). 
Subsampling technique introduced by \texttt{word2vec} \cite{mikolov2013distributed} is applied to accelerate the training process. 
The learning rate is set to 0.025. 
%Dropout is annealed from 0.5 to 0.0 within the first epoch. 
The embedding dimension is 300.
We initialize $Q$ and $V$ as zeros, and $P$ and $U$ from uniform distribution $[-\sqrt{1/100}, \sqrt{1/100}]$ such that each embedding has unit length in expectation \cite{lei2015molding}.
%For the negative sampling distribution in (\ref{w2v}), we use $1/3$ unigram of a word for each of its sense to compute the negative sampling probability using the 3/4rd power trick in \citet{mikolov2013distributed}.
Our model uses 25 negative senses for negative sampling in (\ref{w2v}). We use $\epsilon = 5\%$ for $\epsilon$-Greedy sense selection strategy

In optimization, we conduct mini-batch training with 2048 batch size using the following procedure: 1) select senses in the batch; 2) optimize $U,V$ using stochastic training within the batch for efficiency; 3) optimize $P,Q$ using mini-batch training for robustness. 

%\subsection{Configuration}

%with a 2048 batch size to stabilize the training process, and stochastic gradient ascent
%set the batch size as 2048, and conduct stochastic gradient ascent training (within the batch) on equation (\ref{w2v}) to accelerate training and then use mini-batch gradient descent training on equation (\ref{Qlearn}) to stabilize the sense selection module. 
%To further stabilize the reward signal for the sense selection module, we only use the collocated sense $\sigma(U^T_{z_{ik}}, V_{z_{jl}})$ in (\ref{w2v}) to approximate the collocation likelihood $Co(z_{ik},z_{jl}\mid U,V)$ for Q-learning in (\ref{Qlearn}). 

\subsection{Experiment 1: Contextual Word Similarity}

%\begin{table}
%\small
%\label{SCWS}
%\renewcommand{\arraystretch}{1.1}
%\label{rankcorrelation}
%\vspace{2mm}
%\centering
%\begin{tabular}{| l | c | c | c |}
%\hline
%Method & localSim & avgSim & avgSimC\\
%\hline\hline
%\newcite{HuangEtAl2012} & 26.1 & 62.8 & 65.7\\
%\newcite{neelakantan2015efficient} & 59.8 & 67.3 & 69.1\\
%\newcite{tian2014probabilistic} & 63.6 & - & 65.4\\
%\newcite{li2015multi} & 66.4 & - & -\\
%\newcite{Qiu2016ContextDependentSE} & 64.9 & - & 66.1\\
%Proposed: DRL-Sense & \textbf{66.6} & 64.3 & 65.2\\
%\hline
%\end{tabular}
%\caption{Spearman's rank correlation $\rho$ x100 on the SCWS dataset}
%\end{table}

To evaluate the quality of the learned sense embeddings, we compute the similarity score between each word pair given their respective local contexts and compare with the human-judged score using Stanford's Contextual Word Similarities (SCWS) dataset \cite{HuangEtAl2012}.
Specifically, given a list of word pairs with corresponding contexts, $S = \{(w_i, \bar{C_t}, w_j, \bar{C_{t'}})\}$, we calculate the Spearman's rank correlation $\rho$ between human-judged similarity and model similarity estimations\footnote{For example, human-judged similarity between ``... east \textbf{bank} of the Des Moines River ...'' and ``... basis of all \textbf{money} laundering ...'' is 2.5 out of 10.0 in SCWS dataset~\cite{HuangEtAl2012}.}.
Two major contextual similarity estimations are introduced by \citet{reisinger2010multi}: AvgSimC and MaxSimC.
AvgSimC is a \emph{soft} measurement that addresses the contextual information with a probability estimation:
%\vspace{-4pt}
\begin{align}
%\vspace{-9pt}
\text{AvgSimC} & (w_i, \bar{C_t}, w_j, \bar{C_{t'}}) = \nonumber\\
\sum_{k=1}^{|Z_i|} & \sum_{l=1}^{|Z_j|} \pi(z_{ik}|\bar{C_t}) \pi(z_{jl}|\bar{C_{t'}}) d(z_{ik}, z_{jl}),\nonumber
\end{align}
where $d(z_{ik}, z_{jl})$ refers to the cosine similarity between $U_{z_{ik}}$ and $U_{z_{jl}}$.
AvgSimC weights the similarity measurement of each sense pair $z_{ik}$ and $z_{jl}$ by their probability estimations.
On the other hand, MaxSimC is a \emph{hard} measurement that only considers the most probable senses:
%\vspace{-3pt}
\begin{align*}
\text{MaxSimC}&(w_i, \bar{C_t}, w_j, \bar{C_{t'}}) = d(z_{ik}, z_{jl}),\\
z_{ik}=&\argmax_{z_{ik'}}\pi(z_{ik'}|\bar{C_t}),\\
z_{jl}=&\argmax_{z_{jl'}} \pi(z_{jl'}|\bar{C_{t'}}).
\end{align*}
%\vspace{-3mm}

\begin{table}
%\small
%\renewcommand{\arraystretch}{1.1}
\centering
\begin{tabular}{l c c c}
\hline
\bf Method & \bf \small{MaxSimC} & \bf \small{AvgSimC}\\
\hline\hline
\newcite{HuangEtAl2012} & 26.1 & 65.7\\
\newcite{neelakantan2015efficient} & 60.1 & \bf 69.3\\
\newcite{tian2014probabilistic} & 63.6 & 65.4\\
\newcite{li2015multi} & 66.6 & 66.8\\
\newcite{bartunov2015breaking} & 53.8 & 61.2 \\
\newcite{Qiu2016ContextDependentSE} & 64.9 & 66.1\\
\hline
MUSE-Policy &  66.1 & 67.4\\
MUSE-Greedy & 66.3 & 68.3\\
MUSE-$\epsilon$-Greedy & 67.4$^\dag$ & 68.6\\
MUSE-Boltzmann & \bf 67.9$^\dag$ & 68.7\\
%MUSE - Thompson & 65.9 & 67.9\\
\hline
\end{tabular}
\vspace{-2mm}
\caption{Spearman's rank correlation $\rho$ x100 on the SCWS dataset. $^\dag$ denotes superior performance to all unsupervised competitors.}
%\vspace{-2mm}
\label{tab:scws}
\end{table}

The baselines for comparison include
classic clustering methods \cite{HuangEtAl2012,neelakantan2015efficient}, EM algorithms \cite{tian2014probabilistic,Qiu2016ContextDependentSE,bartunov2015breaking}, and Chinese Restaurant Process \cite{li2015multi}\footnote{We run \citet{li2015multi}'s released code on our corpus for fair comparison.}, 
where all approaches are trained on the same corpus except \citet{Qiu2016ContextDependentSE} used more recent Wikipedia dumps.
The embedding sizes of all baselines are 300, except 50 in \citet{HuangEtAl2012}.
For every competitor with multiple settings, we report the best performance in each similarity measurement setting and show in Table~\ref{tab:scws}.

\begin{table}[t]
%\vspace{2mm}
\centering
\begin{tabular}{ l  c  c  c }  
\hline
\bf Method & \bf\small ESL-50 & \small \bf RD-300 & \small \bf TOEFL-80\\
\hline\hline
\multicolumn{4}{l}{\textit{1) Conventional Word Embedding}}\\
Global Context & 47.73 & 45.07 & 60.87\\
Skip-Gram & 52.08 & 55.66 & 66.67\\
\hline
\multicolumn{4}{l}{\textit{2) Word Sense Disambiguation}}\\
IMS+SG & 41.67 & 53.77 & 66.67\\
\hline
\multicolumn{4}{l}{\textit{3) Unsupervised Sense Embeddings}}\\
EM & 27.08 & 33.96 & 40.00\\
MSSG & 57.14 & 58.93 & 78.26\\
CRP & 50.00 & 55.36 & 82.61\\
\hline
MUSE-\small{Policy} & 52.38 & 51.79 & 79.71\\
MUSE-\small{Greedy} & 57.14 & 58.93 & 79.71\\
MUSE-\small{$\epsilon$-Greedy} & 61.90$^\dag$ & 62.50$^\dag$ & 84.06$^\dag$\\
MUSE-\small{Boltzmann} & \bf 64.29$^\dag$ & \bf 66.07$^\dag$ & \bf 88.41$^\dag$\\
\hline\hline
\multicolumn{4}{l}{\textit{4) Supervised Sense Embeddings}}\\
Retro-GC & 63.64 & 66.20 & 71.01\\
Retro-SG & 56.25 & 65.09 & 73.33\\
\hline
\end{tabular}
\vspace{-2mm}
\caption{Accuracy on synonym selection. $^\dag$ denotes superior performance to all unsupervised competitors.}
%\vspace{-2mm}
\label{tab:synonym}
\end{table}	

Our MUSE model achieves the state-of-the-art performance on MaxSimC, demonstrating superior quality on independent sense embeddings.
On the other hand, MUSE achieves comparable performance with the best competitor in terms of AvgSimC (68.7 vs. 69.3), while MUSE outperforms the same competitor significantly in terms of MaxSimC (67.9 vs. 60.1). 
The results demonstrate not only the high quality of sense representations but also accurate sense selection.

From the application perspective, MaxSimC refers to a typical scenario using single embedding per word, while AvgSimC employs multiple sense vectors simultaneously per word, which not only brings computational overhead but changes existing neural architecture for NLP.
%using multiple sense vectors for a word simultaneously like AvgSimC may bring computational overhead over conventional single word embedding scenarios like MaxSimC, and would change existing neural network architecture with a single embedding per word.
Hence, we argue that MaxSimC better characterize practical usage of a sense representation system than AvgSimC. 

Among various learning methods for MUSE, policy gradient performs worst, echoing our argument in \cref{sec:pg}. On the other hand, the superior performance of Boltzmann sampling and $\epsilon$-Greedy over Greedy selection demonstrates the effectiveness of exploration.

Finally, replacing $\bar{\mathcal{L}}(\cdot)$ with $\hat{\mathcal{L}}(\cdot)$ as the reward signal yields $2.3$ times speedup for MUSE-$\epsilon$-Greedy and $1.3$ times speedup for MUSE-Boltzmann to reach 67.0 in MaxSimC, which demonstrates the efficacy of proposed approximation $\hat{\mathcal{L}}(\cdot)$ over typical $\bar{\mathcal{L}}(\cdot)$ in terms of convergence.

\subsection{Experiment 2: Synonym Selection}

We further evaluate our model on synonym selection using multi-sense word representations~\cite{jauhar2015ontologically}.
%A supervised sense embedding system to perform evaluation on synonym selection. 
Three standard synonym selection datasets, ESL-50 \cite{turney2001mining}, RD-300 \cite{jarmasz2004roget}, and TOEFL-80 \cite{landauer1997solution}, are performed.
In the datasets, each question consists of a question word $w_Q$ and four answer candidates $\{w_A, w_B, w_C, w_D\}$, and the goal is to select the most semantically synonymous choice among the four candidates.
For example, in the TOEFL-80 dataset, a question shows $\{$(Q) enormously, (A) appropriately, (B) uniquely, (C) tremendously, (D) decidedly$\}$, and the answer is (C). 
For multi-sense representations system, it selects the synonym of the question word $w_Q$ using the maximum sense-level cosine similarity as a proxy of the semantic similarity \cite{jauhar2015ontologically}.
%For pairs of words $(w_Q, w_T), T\in \{A,B,C,D\}$, the maximum sense-level cosine similarity is computed as
%\begin{equation}
%\text{MaxSim}(w_Q, w_T) =\max_{k\in Z_Q, l\in Z_T} d(z_{Qk}, z_{Tl}). \nonumber
%\end{equation}

Our model is compared with the following baselines: 
1) conventional word embeddings: global context vectors \cite{HuangEtAl2012} and skip-gram \cite{mikolov2013distributed};
2) applying supervised word sense disambiguation using the IMS system  and then applying skip-gram on disambiguated corpus (IMS+SG)~\cite{zhong2010makes};
3) unsupervised sense embeddings: EM algorithm \cite{jauhar2015ontologically}, multi-sense skip-gram (MSSG) \cite{neelakantan2015efficient}, Chinese restaurant process (CRP) \cite{li2015multi}, and the MUSE models;
4) supervised sense embeddings with WordNet~\cite{miller1995wordnet}:
retrofitting global context vectors (Retro-GC) and retrofitting skip-gram (Retro-SG) \cite{jauhar2015ontologically}.

Among unsupervised sense embedding approaches, CRP and MSSG refer to the baselines with highest MaxSimC and AvgSimC in Table~\ref{tab:scws} respectively.
%We note that, for competitors in unsupervised sense embeddings, CRP refers to the best competitor in terms of MaxSimC and MSSG refers to the best competitor in terms of AvgSimC in the SCWS dataset. 
Here we report the setting for baselines based on the best average performance in this task.
We also show the performance of supervised sense embeddings as an upperbound of unsupervised methods
%Note that we do not compare with competitors in 4) on SCWS dataset, because these methods do not include a sense selector, and thus cannot be evaluated using MaxSimC and AvgSimC.
%supervised sense embeddings can be treated as an upperbound of unsupervised methods 
due to the usage of additional supervised information from WordNet.

\begin{table*}[t]
%\small
%\renewcommand{\arraystretch}{1.25}
\centering
\begin{tabular}{l l }
\hline
\bf Context & \bf k-NN Senses\\
%\rowcolor{gray}
%apple(3)/N & apple blackberry amstrad guava apricot\\
\hline\hline
\small $\cdots$ braves finish the season in \textbf{tie} with the los angeles dodgers $\cdots$ & \small scoreless otl shootout 6-6 hingis 3-3 7-7 0-0\\
\small $\cdots$ his later years proudly wore \textbf{tie} with the chinese characters for $\cdots$ & \small pants trousers shirt juventus blazer socks anfield \\
\hline
\small $\cdots$ of the mulberry or the \textbf{blackberry} and minos sent him to $\cdots$ & \small cranberries maple vaccinium apricot apple \\%  blackberries blueberry milk prunus\\
\small $\cdots$ of the large number of \textbf{blackberry} users in the us federal $\cdots$ & \small smartphones sap microsoft ipv6 smartphone\\% linux-based pda messaging symbian\\
\hline
\small $\cdots$ shells and/or high explosive squash \textbf{head} hesh and/or anti-tank $\cdots$ & \small venter thorax neck spear millimeters fusiform \\% beachy head shaved maldives whale\\
\small $\cdots$ head was shaven to prevent \textbf{head} lice serious threat back then $\cdots$ & \small shaved thatcher loki thorax mao luthor chest \\% pressure
\small $\cdots$ appoint john pope republican as \textbf{head} of the new army of $\cdots$ & \small multi-party appoints unicameral beria appointed \\ %minister-president cabinet thatcher elect coach shaved\\
\hline
\end{tabular}
\vspace{-2mm}
\caption{%Words with different contexts occurring in the training corpus. 
Different word senses are selected by MUSE according to different contexts. The respective k-NN (sorted by collocation likelihood) senses are shown to indicate respective semantic meanings.} %The existence (Y/N) indicates whether the sense representation has been selected at least once throughout the training procedure.}
\label{tab:qual}
\end{table*}

The results are shown in Table~\ref{tab:synonym}, where our MUSE-$\epsilon$-Greedy and MUSE-Boltzmann significantly outperform all unsupervised sense embeddings methods, echoing the superior quality of our sense vectors in last section.
MUSE-Boltzmann also outperforms the supervised sense embeddings except 1 setting without any supervised signal during training.
Finally, the MUSE methods with proper exploration outperform all unsupervised baselines consistently, demonstrating the importance of exploration.
%Except for retrofitting methods, our MUSE model significantly outperforms all baselines, echoing the superior quality of our sense vectors in last section.
%In addition, our model also outperforms Retro-SG in two out of three datasets, although Retro-SG uses the supervised knowledge from WordNet, showing that our model can obtain better semantic representation in an unsupervised fashion than supervised method in some settings.

\subsection{Qualitative Analysis}

We further conduct qualitative analysis to check the semantic meanings of different senses learned by MUSE with k-nearest neighbors (k-NN) using sense representations. 
In addition, we provide contexts in the training corpus where the sense will be selected to validate the sense selection module.
Table \ref{tab:qual} shows the results.
The learned sense embeddings of the words ``tie'', ``blackberry'', and ``head'' clearly correspond to correct senses under different contexts.

Since we address an unsupervised setting that learns sense embeddings from unannotated corpus, the discovered senses highly depend on the training corpus. From our manual inspection, it is common for our model to discover only two senses in a word, like “tie” and “blackberry”. However, we maintain our effort in developing unsupervised sense embeddings learning methods in this work, and the number of discovered sense is not a focus.

\section{Conclusion}

%This paper proposes a novel modularization framework, which supports joint training on a sense representation module and a sense selection module. 
This paper proposes a novel modularized framework for unsupervised sense representation learning, which supports not only the flexible design of modular tasks but also joint optimization among modules.
%Without the need to design several mechanisms within a single model, the modular framework supports flexible design on each module to optimize corresponding tasks.
%The resulting MUSE model can exhibit several nice property simultaneously due to mechanism-wise optimization for each module.
%Through modularization,
The proposed model is the first work that implements purely sense-level representation learning with linear-time sense selection, and 
%The experiments show that our MUSE model 
achieves the state-of-the-art performance on benchmark contextual word similarity and synonym selection tasks.
%synonym selection and on benchmark contextual word similarity task in terms of MaxSimC. 
In the future, we plan to investigate reinforcement learning methods to incorporate multi-sense word representations for downstream NLP tasks.

%\clearpage

%\section{Implementation Details}
%\label{parameter_setting}

\section*{Acknowledgements}
We would like to thank reviewers for their insightful comments on the paper.
The authors are supported by the Ministry of Science and Technology of Taiwan under the contract number 105-2218-E-002-033, Institute for Information Industry, and MediaTek Inc..

\bibliography{emnlp2017}
\bibliographystyle{emnlp_natbib}
%\if 0

%\clearpage
\appendix

\section{Doubly Stochastic Gradient}
\label{doubly_sg}

%For clarity purpose, we abbreviate the collocation log likelihood $\log Co(ik,jl|U,V)$ as $\mathcal{L}(\cdot)$ in this section.
%For clarity purpose, we abbreviate the collocation log likelihood $\log \mathcal{L}(z_{jl}|z_{ik})$ as $\log \mathcal{L}(\cdot)$ in this section.
To derive doubly stochastic gradient for equation (\ref{policy_grad}), we first denote (\ref{policy_grad}) as $J(\Theta)$ with $\Theta=\{P, Q\}$ and resolve the expectation form as:
\begin{align*}
J(\theta)& = \mathbb{E}_{z_{ik} \sim \pi(\cdot \mid \bar{C_t})}[\log \bar{\mathcal{L}}(z_{jl} \mid z_{ik})]\\
& = \sum_k \pi(z_{ik}\mid \bar{C_t})\log \bar{\mathcal{L}}(z_{jl}|z_{ik}).
\end{align*}
Denote $\Theta=\{P, Q\}$ as the parameter set for policy $\pi$. The gradient with respect to $\Theta$ should be:
\begin{align*}
& \frac{\partial J(\theta)}{\partial \Theta} = \frac{\partial}{\partial \Theta} \sum_k \pi(z_{ik}\mid \bar{C_t})\log \bar{\mathcal{L}}(z_{jl}|z_{ik})\\
%& = \frac{\partial}{\partial \Theta} \sum_k \pi(z_{ik}\mid \bar{C_t})\log \bar{\mathcal{L}}(z_{jl}|z_{ik}) \\
& = \sum_k \log \bar{\mathcal{L}}(z_{jl}|z_{ik}) \frac{\partial \pi(z_{ik}\mid \bar{C_t})}{\partial \Theta}\\
& = \sum_k \log \bar{\mathcal{L}}(z_{jl}|z_{ik}) (\frac{\partial \log \pi(z_{ik}\mid \bar{C_t})}{\partial \Theta})(\pi(z_{ik}\mid \bar{C_t}))\\
& = \mathbb{E}_{z_{ik} \sim \pi(\cdot \mid \bar{C_t})}[\log \bar{\mathcal{L}}(z_{jl} \mid z_{ik}) \frac{\partial \log \pi(z_{ik}\mid \bar{C_t})}{\partial \Theta}]
\end{align*}
Accordingly, if we conduct typical stochastic gradient ascent training on $J(\Theta)$ with respect to $\Theta$ from samples $z_{ik}$ with a learning rate $\eta$, the update formula will be:
\begin{align*}
\Theta = \Theta + \eta \log \bar{\mathcal{L}}(z_{jl} \mid z_{ik}) \frac{\partial \log \pi(z_{ik}\mid \bar{C_t})}{\partial \Theta}.
\end{align*}
However, the collocation log likelihood should always be non-positive: $\log \bar{\mathcal{L}}(z_{jl} \mid z_{ik})\leq 0$. Therefore, as long as the collocation log likelihood $\log \bar{\mathcal{L}}(z_{jl} \mid z_{ik})$ is negative, the update formula is to minimize the likelihood of choosing $z_{ik}$, despite the fact that $z_{ik}$ may be good choices. On the other hand, if the log likelihood reaches $0$, according to (\ref{w2v}), it indicates:
\begin{align*}
& \log \bar{\mathcal{L}}(z_{jl} \mid z_{ik}) = 0 \Rightarrow\;\;\; \bar{\mathcal{L}}(z_{jl} \mid z_{ik}) = 1\\
%\Rightarrow\;\;\; & \bar{\mathcal{L}}(z_{jl} \mid z_{ik}) = 1\\
\Rightarrow\;\;\; & U^T_{z_{ik}} V_{z_{jl}} \rightarrow \infty,\;\; U^T_{z_{ik}} V_{z_{uv}} \rightarrow \infty,\;\;\forall {z_{uv}},
\end{align*}
which leads to computational overflow from an infinity value.

%\fi

\end{document}